\documentclass[11pt]{article}

\usepackage[preprint]{acl}

\usepackage{times}
\usepackage{latexsym}
\usepackage[T1]{fontenc}
\usepackage[utf8]{inputenc}
\usepackage{microtype}
\usepackage{graphicx}
\usepackage{float}
\usepackage{booktabs}
\usepackage{amsmath}
\usepackage{amssymb}
\usepackage{xcolor}
\usepackage{subcaption}

\title{Same Question, Different Source, Different Answer:\\
       Auditing Source-Dependence in
       Medical Multi-Source RAG}

\author{Yubo Li, Rema Padman, Ramayya Krishnan\\
Carnegie Mellon University\\
\{yubol, rpadman, rk2x\}@andrew.cmu.edu}

\begin{document}
\maketitle

\begin{abstract}
A retrieval-augmented generation (RAG) system deployed over a multi-author institutional corpus can give a different answer to the same question depending on which source it retrieves --- a failure mode the dominant single-gold-answer paradigm cannot diagnose. We argue that \emph{source-dependence} is a missing axis of NLP evaluation, and that auditing it means shifting the unit of evaluation from answer correctness to the \emph{inter-source relationship}. We make this concrete in transplant patient education, where institutional sources demonstrably disagree, releasing three artefacts: \textbf{TransplantQA}, a benchmark of real patient questions, each answered by grounding generation in multiple institutional handbooks as candidate sources; \textbf{HERO-QA}, a hierarchical retrieval strategy that grounds and audits each answer; and a structured-output judge that scores inter-source relationships on a validated 5-label taxonomy. At scale, better retrieval reveals far more disagreement than prior estimates suggested --- understating its \emph{prevalence}, not its intensity. The framework is domain-agnostic and transfers to legal and educational RAG: measuring source-dependence is a responsibility for deployed multi-source NLP generally.
\end{abstract}

\section{Introduction}
\label{sec:intro}

A patient three months past a heart transplant types a question into an institutional Q\&A system: \emph{``When can I travel internationally again?''}\footnote{Adapted from a real patient post on a transplant forum included in our benchmark.} Behind the system, an RAG pipeline retrieves passages from the patient-education handbook of the institution that performed the surgery. The answer is grounded, cited, and confidently delivered. Had the same query been grounded in a peer institution's handbook, the recommended waiting period might have been three, six, or twelve months --- with identical confidence and fluency, and no indication that the guidance is institution-specific rather than universal.

This kind of \emph{inter-source heterogeneity} is endemic to medical RAG. Patient-facing institutional documents reflect local protocols, editorial choices, and decades of accumulated risk-management caution; they are not interchangeable. Yet the dominant benchmarks for medical question answering --- MedQA \citep{jin2021medqa}, MedMCQA \citep{pal2022medmcqa}, PubMedQA \citep{jin2019pubmedqa}, BioASQ \citep{tsatsaronis2015bioasq} --- assume one correct answer per question and cannot diagnose whether the answer a patient sees is contingent on which document the retriever happened to return.

We argue this exposes a missing axis of NLP evaluation. As RAG becomes deployed infrastructure over multi-author institutional corpora --- in medicine, but equally in law and education --- the field needs to measure \emph{source-dependence}: whether the answer a user receives is contingent on which source the retriever happened to return. We frame this as a new mission for evaluation research, and operationalise it by shifting the unit of analysis from single-answer correctness to \emph{inter-source relationship}: given the same question, what is the structured relationship between the answer a generator produces when grounded in document $A$ versus document $B$? This paper makes four contributions toward that shift, using transplant patient education as a case study in which institutional sources demonstrably disagree.

\begin{enumerate}\itemsep -1pt
  \item \textbf{An evaluation-paradigm argument} (§\ref{sec:intro}, §\ref{sec:discussion}): the single-gold-answer paradigm cannot diagnose source-dependence, the dominant failure mode of deployed multi-source RAG; closing the gap requires evaluating the inter-source relationship, not refining single-gold benchmarks.

  \item \textbf{TransplantQA} (§\ref{sec:benchmark}): a benchmark operationalising this shift --- 1{,}115 real patient questions, each answered by grounding generation in 102 transplant patient-education handbooks (the candidate sources) from 23 U.S.\ centers across five organ types, partitioned into a \emph{general} subset (answered by every handbook) and an \emph{organ-specific} subset, enabling both full-corpus and stratified inter-source comparison.

  \item \textbf{HERO-QA} (§\ref{sec:retrieval_gen}): a hierarchical evidence retrieval and orchestration strategy for handbook-grounded clinical QA, using full-document context for short handbooks (eliminating retrieval-miss failures) and section-aware hierarchical retrieval with reranking for longer ones, with explicit retrieval metadata for grounding audit.

  \item \textbf{Empirical characterization at scale} (§\ref{sec:characterization}): the full output of a production run over the benchmark (48{,}056 grounded answers, 5{,}730{,}465 pairwise comparisons), released for reuse. The inter-source relationship is measured by a structured-output judge (the evaluation instrument; §\ref{sec:judge}) validated against human annotators at $\kappa=0.842$ (§\ref{sec:validation}).
\end{enumerate}

Our characterization also yields a methodological observation: comparing the reference run against an earlier 14B run with a lower-capacity retriever, the average handbook absence rate drops 13.6\,pp while per-pair divergence is essentially unchanged (\S\ref{sec:14B_vs_32B}) --- prior estimates understated the \emph{prevalence} of disagreement, not its \emph{intensity}. Crucially, the framework is not medicine-specific: legal RAG (retrieving over federal/state/circuit precedent) and educational RAG (retrieving over state-stratified curriculum standards) deploy over the same kind of multi-source corpora and inherit the same blind spot, and the three components --- multi-source benchmark, inter-source taxonomy, structured-output judge --- transfer directly to both (\S\ref{sec:discussion}). Measuring source-dependence is thus a mission for deployed multi-source NLP broadly, not a medical-domain convenience.

\section{Related Work}
\label{sec:related}

\paragraph{Medical QA benchmarks.} Medical-QA evaluation treats QA as single-best-answer prediction: MedQA \citep{jin2021medqa}, MedMCQA \citep{pal2022medmcqa}, PubMedQA \citep{jin2019pubmedqa}, and BioASQ \citep{tsatsaronis2015bioasq} score against curated gold answers, and patient-facing extensions \citep{abacha2019liveqa,zeng2020meddialog,singhal2023medpalm} retain the single-gold assumption. TransplantQA instead makes the \emph{relationship} between answers grounded in different documents the unit of analysis; to our knowledge no prior medical QA benchmark tests inter-source heterogeneity at this scale.

\paragraph{LLM-as-judge and cross-document inconsistency.} LLM-as-judge protocols \citep{zheng2024mtbench,zhu2023judgelm,kim2024prometheus,liu2023geval} typically return a single scalar or label; our judge instead co-emits narrative metadata (\texttt{divergence\_topic}, \texttt{clinical\_significance}), enabling the taxonomy and severity analyses of §\ref{sec:characterization} at essentially unchanged per-pair cost. Separately, contradiction detection via NLI \citep{schuster2022stretching}, factuality decomposition \citep{min2023factscore}, and RAG-hallucination evaluation \citep{niu2024ragtruth} target a binary signal against a reference; we instead treat each answer as faithful to its source and ask whether two sources \emph{themselves} agree, with a 5-label taxonomy that surfaces \textsc{Complementary}/\textsc{Divergent} variation a binary lens misses.

\paragraph{Institutional variation in medicine.} \citet{wennberg1973small} documented small-area variation in clinical practice unexplained by patient characteristics, launching a long literature on clinical-practice variation. Patient-facing educational material is the visible boundary of this institutional variation; TransplantQA provides an NLP-tractable instrument for measuring it.

\section{The TransplantQA Benchmark}
\label{sec:benchmark}

TransplantQA pairs a corpus of patient-education handbooks from U.S.\ transplant centers with a question set drawn from real patient information-seeking behavior, so that an RAG system's answer to any benchmark question can be grounded in (and evaluated against) multiple plausible institutional sources. Unlike single-gold medical QA benchmarks, the unit of analysis in TransplantQA is the inter-source \emph{relationship} between answers grounded in different documents.

\subsection{Handbook Corpus}
\label{sec:corpus}

We collected 102 patient-education handbooks from 23 major U.S.\ solid-organ transplant centers, representing 16 of the 20 largest programs by procedure volume. The corpus spans five organ types --- heart (26), lung (26), kidney (22), liver (17), and pancreas (11) --- and the contributing institutions are geographically distributed across the United States, comprising both large academic medical centers and community-based transplant programs. All documents were obtained as PDFs from institutional websites and patient education portals.

Centers organize patient education differently: some provide separate documents for the pre-transplant phase (evaluation, listing, waiting) and the post-transplant phase (recovery, medications, long-term follow-up), while others issue a single combined handbook. We treat each phase-specific document as a distinct unit, yielding 37 pre-transplant, 39 post-transplant, and 26 combined handbooks. Each is assigned an identifier encoding organ, institution, and care phase (e.g., \texttt{heart\_baylor\_combined}). Table~\ref{tab:handbook_summary} summarizes the corpus.

\begin{table}[t]
\centering
\small
\caption{TransplantQA handbook corpus by organ. \emph{Centers} is the number of distinct contributing institutions.}
\label{tab:handbook_summary}
\begin{tabular}{lrrrrr}
\toprule
 & \textbf{Heart} & \textbf{Kidney} & \textbf{Liver} & \textbf{Lung} & \textbf{Panc.} \\
\midrule
Handbooks       & 26 & 22 & 17 & 26 & 11 \\
Centers         & 17 & 14 & 11 & 15 &  8 \\
Pre             & 10 &  8 &  5 & 10 &  4 \\
Post            & 11 & 10 &  4 & 11 &  3 \\
Combined        &  5 &  4 &  8 &  5 &  4 \\
\bottomrule
\end{tabular}
\end{table}

\subsection{Question Set}
\label{sec:questions}

\begin{figure*}[t]
\centering
\includegraphics[width=0.82\textwidth]{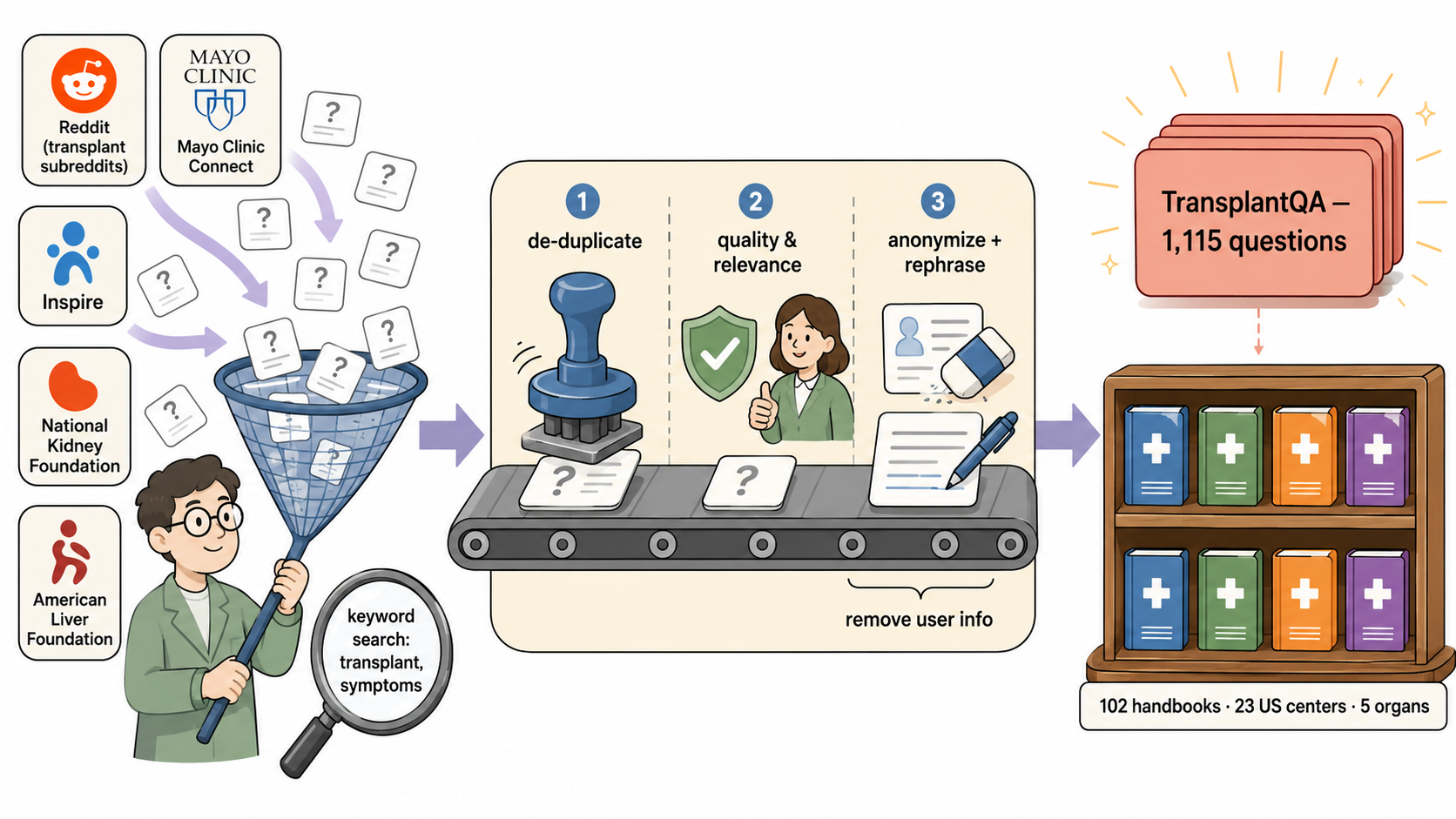}
\caption{TransplantQA construction. Patient questions are harvested from real online transplant communities and platforms (patient forums and social media, patient-advocacy organizations, and institutional Q\&A) via transplant- and symptom-keyword search, then de-duplicated, quality/relevance-checked, and anonymized and rephrased to remove user-identifying information --- yielding 1{,}115 questions (311 general answered by every handbook + 804 organ-specific), paired with 102 patient-education handbooks from 23 U.S.\ centers across five organ types.}
\label{fig:funnel}
\end{figure*}

We curated 1{,}115 patient questions to serve as the evaluation set for cross-center comparison (Figure~\ref{fig:funnel}). Questions were \emph{harvested from real online transplant communities and platforms} --- patient forums and social media (e.g.,\ Reddit transplant subreddits, Mayo Clinic Connect, Inspire), patient-advocacy organizations (National Kidney Foundation, American Liver Foundation), and institutional Q\&A pages --- using transplant- and symptom-keyword search to surface genuine information needs. The 3{,}000+ harvested candidates were then (i)~de-duplicated (cosine $>0.85$ plus manual review), (ii)~double-checked for quality and relevance, and (iii)~\emph{anonymized and rephrased} to strip user-identifying content and make each question self-contained, yielding the released 1{,}115 (mean length 23.6 words). Source breakdown and inclusion criteria are in Appendix~\ref{app:question_sources}.

Each question is annotated with: (i) an \emph{organ-type label} --- heart, kidney, liver, lung, pancreas, or \emph{general}; (ii) one or more clinical topic categories drawn from a 13-topic taxonomy (Appendix~\ref{app:topic_taxonomy}); and (iii) fine-grained sub-topic tags (43 unique). Questions are multi-labeled to reflect cross-cutting concerns.

\paragraph{General vs.\ organ-specific split.}
A central design choice is the partition of the question set into a \emph{general} subset (311 questions, 27.9\%) and an \emph{organ-specific} subset (804 questions across five organ types). General questions address topics relevant to all transplant recipients --- immunosuppressant side effects, reproductive health, mental health --- and are answered by \emph{every} handbook in the corpus, producing $\binom{102}{2} = 5{,}151$ pairwise comparisons per question. Organ-specific questions are answered only by handbooks of the matching organ type, producing $\binom{N_o}{2}$ comparisons where $N_o\in\{11,17,22,26,26\}$. The two subsets together support both full-corpus and stratified inter-source analyses.

\subsection{Anonymization and Release}
\label{sec:anonymization}

Because questions are harvested from public forums and social media, every released question was anonymized and rephrased to remove any user-identifying content from the original post (Appendix~\ref{app:question_sources}); the released benchmark also uses anonymized handbook identifiers. Center names in handbook IDs are retained because transplant centers are public institutions and the analyses we enable are explicitly cross-institutional. Release-location metadata is anonymized for review; the planned release package includes the benchmark, the raw handbook-extraction output, the question annotations, and the full pairwise-comparison outputs. Original PDFs are not redistributed but are listed by URL for independent retrieval. Appendix~\ref{app:datacard} provides a Datasheet-style data card~\citep{gebru2021datasheets}.

\section{Pipeline Architecture}
\label{sec:hero}

Our pipeline is a three-stage process that takes the benchmark question set and the handbook corpus as input and produces, for every benchmark question, a structured matrix of pairwise inter-handbook relationships. It runs on open-weight LLMs (Qwen3-32B for both generation and judging in our reference run) and is designed for resumable execution on heterogeneous SLURM clusters. The methodological core of this section is \emph{HERO-QA}, the hierarchical evidence-retrieval strategy used in Stage~2 (§\ref{sec:retrieval_gen}, Figure~\ref{fig:hero}); the structured pairwise judge in Stage~3 (§\ref{sec:judge}) is the measurement instrument that operationalises the inter-source evaluation.

\subsection{Stage 1: Structured Extraction}
\label{sec:extraction}

Raw PDF handbooks are converted to structured JSON using LlamaParse~\citep{llamaindex_llamaparse}, preserving section headings, paragraph boundaries, and page metadata. The per-handbook output contains organ type, institution, care phase, source path, full text, and a section list with headings, body text, and page numbers. This structure enables section-aware chunking in Stage~2. Extraction is idempotent.

\subsection{Stage 2: HERO-QA Retrieval-Augmented Generation}
\label{sec:retrieval_gen}

\begin{figure*}[t]
\centering
\includegraphics[width=0.86\textwidth]{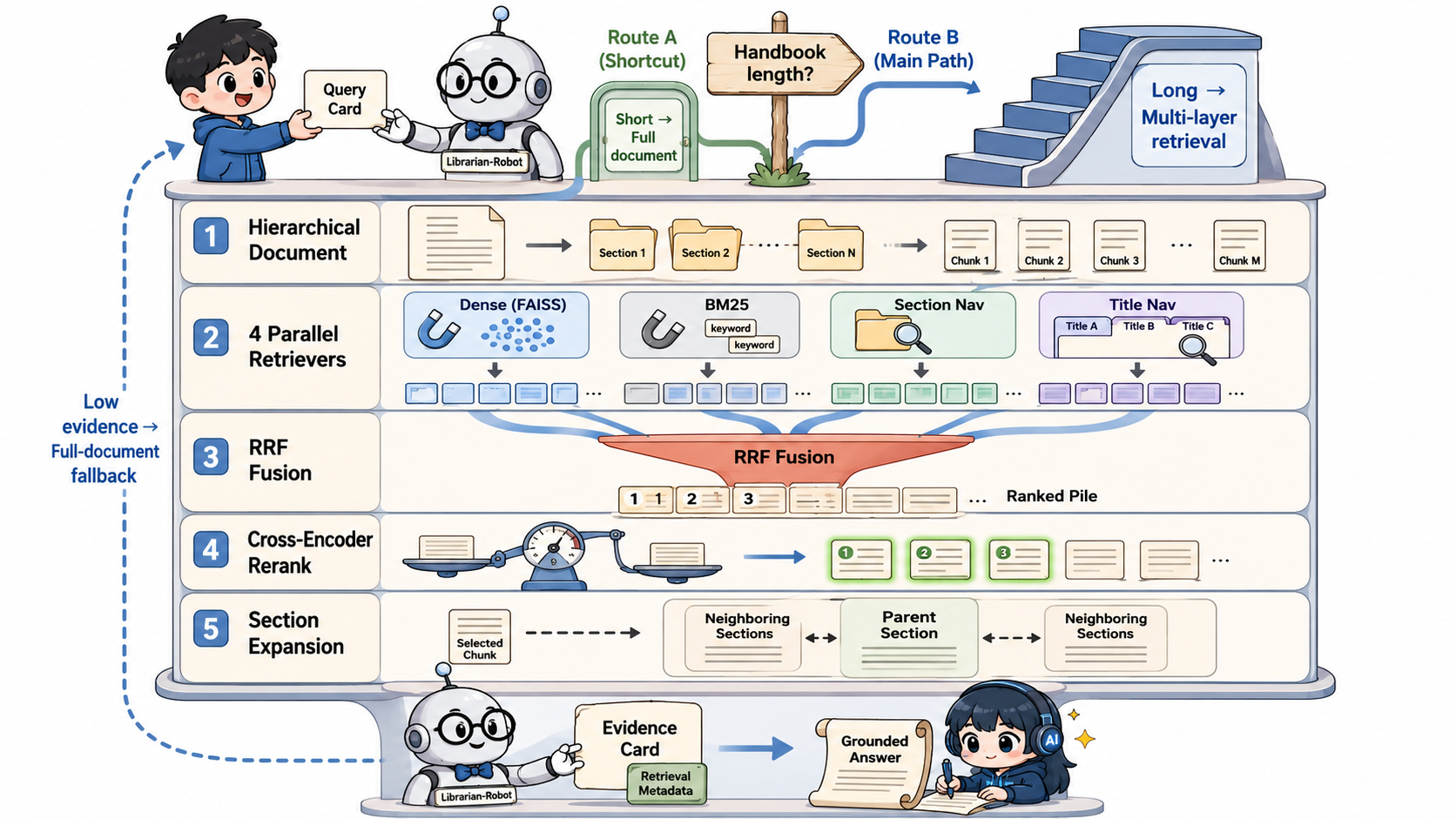}
\caption{HERO-QA: a multi-layer retrieval system. A query is routed by handbook length: short handbooks bypass retrieval and use full-document context (Route~A); long handbooks descend through a hierarchical document model (document $\rightarrow$ sections $\rightarrow$ child chunks), four parallel first-stage retrievers (dense FAISS, child BM25, section-body navigation, title navigation), RRF fusion, cross-encoder reranking, and parent-section expansion. The top evidence grounds Qwen3-32B generation; retrieval metadata is retained for audit, and a low-evidence signal triggers full-document fallback.}
\label{fig:hero}
\end{figure*}

\textbf{HERO-QA} (\textit{Hierarchical Evidence Retrieval and Orchestration for Handbook-grounded clinical QA}) is the retrieval strategy used in Stage~2 (Figure~\ref{fig:hero}). It is a recall-first \emph{multi-layer} retrieval system designed for the institutional-handbook setting, in which a query descends through a length-routing gate, a hierarchical document model, four parallel first-stage retrievers, rank fusion, cross-encoder reranking, and parent-section expansion. Throughout, HERO-QA exposes retrieval metadata (which mode produced the context, which sections were touched) so downstream evaluation can audit whether an answer was grounded in full-document or retrieved evidence.

\textit{Routing and document model (Layers 0--1).} Short handbooks (full text $\leq 80$k chars) are passed in full and retrieval is skipped, eliminating retrieval-miss for short documents. Longer handbooks are decomposed into \emph{parent sections} (preserving headings/pages) and overlapping \emph{child chunks} (160 words, 32-word overlap, each prefixed with its parent heading); this document$\rightarrow$section$\rightarrow$chunk hierarchy is the substrate for retrieval and expansion.

\textit{Four parallel retrievers + fusion + rerank (Layers 2--4).} Against the expanded query, HERO-QA runs four first-stage retrievers: dense child-chunk retrieval (FAISS~\citep{douze2025faiss} with \texttt{BAAI/bge-large-en-v1.5}~\citep{xiao2024c}), sparse child-chunk BM25~\citep{robertson2009probabilistic}, \emph{section-body navigation} (BM25 over section text, hits mapped to child chunks), and \emph{title navigation} (BM25 over section headings, catching topic matches when body wording differs). The four rankings are combined by Reciprocal Rank Fusion ($k_{\mathrm{RRF}}\!=\!60$~\citep{cormack2009reciprocal}; navigation signals down-weighted) and reranked with a MiniLM cross-encoder~\citep{wang2020minilm}.

\textit{Parent-section expansion (Layer 5).} Top child chunks are expanded back to their parent sections plus immediate neighbours, so the generator receives coherent section-level context; the top-5 expanded passages form the evidence. An evidence-sufficiency check triggers full-document fallback when retrieved evidence is weak.

\textit{Answer generation.} For each (question, handbook) pair the retrieved passages are supplied to Qwen3-32B at temperature~0 with a fixed prompt (Appendix~\ref{app:gen_prompt}) instructing the model to (a)~rely exclusively on the provided context, (b)~return a standardized \texttt{NOT ADDRESSED} prefix when the handbook contains no relevant information rather than fabricate, and (c)~cite the supporting section heading when one exists. The stage produces 48{,}056 grounded answers in the reference run.

\subsection{Stage 3: Structured Pairwise Judgment}
\label{sec:judge}

\paragraph{Absence pre-screen.} Each answer is first screened for absence: a fast heuristic checks for the canonical \texttt{NOT ADDRESSED} prefix, and answers that escape the heuristic are passed to a binary classifier (also Qwen3-32B) using a structured YES/NO prompt. Absence is cached per (handbook, question) pair, so each handbook is screened once across all comparisons it participates in. Any pair containing at least one absent answer is immediately assigned the \textsc{Absent} label, skipping the comparison call.

\paragraph{Five-label taxonomy.} For every pair of non-absent answers, the judge classifies their relationship into one of five categories with operational definitions (Table~\ref{tab:taxonomy}). The taxonomy is designed to be (a)~clinically interpretable, (b)~jointly exhaustive over the relationships we observed during pilot annotation, and (c)~ordered along a coverage--agreement axis from no information (\textsc{Absent}) through full alignment (\textsc{Consistent}), additive but compatible content (\textsc{Complementary}), substantive but bounded disagreement (\textsc{Divergent}), to outright opposition (\textsc{Contradictory}).

\begin{table*}[t]
\centering
\small
\caption{Five-label taxonomy for pairwise comparison of center-specific answers. Examples are drawn from the released benchmark.}
\label{tab:taxonomy}
\renewcommand{\arraystretch}{1.15}
\begin{tabular}{@{}l p{5.5cm} p{7cm}@{}}
\toprule
\textbf{Label} & \textbf{Definition} & \textbf{Example} \\
\midrule
\textsc{Absent} & One or both answers indicate the handbook does not address the topic. & Center~A provides dietary guidance; Center~B's handbook has no relevant section. \\
\textsc{Consistent} & Same clinical recommendation, no meaningful informational difference. & Both centers advise avoiding grapefruit due to tacrolimus interactions. \\
\textsc{Complementary} & Clinically compatible but differing in detail or scope. & Center~A lists side effects; Center~B additionally describes management strategies. \\
\textsc{Divergent} & Substantive, clinically meaningful difference (different thresholds, timelines, or recommended actions). & Center~A recommends exercise at 6 weeks post-transplant; Center~B at 8--12 weeks. \\
\textsc{Contradictory} & Directly opposing clinical guidance. & Center~A allows ABO-incompatible live donors; Center~B states they cannot proceed. \\
\bottomrule
\end{tabular}
\end{table*}

\paragraph{Structured output beyond the label.} A standard LLM-as-judge protocol would return only the classification. Our judge instead returns a structured JSON record per pair containing five fields:
\begin{enumerate}\itemsep -2pt
  \item \texttt{classification} --- one of the five labels;
  \item \texttt{reasoning} --- a 2--3 sentence clinical justification;
  \item \texttt{divergence\_topic} --- a short noun phrase naming the \emph{locus} of disagreement (emitted only when \texttt{classification} $\not\in \{\textsc{Consistent}, \textsc{Absent}\}$);
  \item \texttt{clinical\_significance} $\in\{\mathrm{low},\mathrm{medium},\mathrm{high}\}$ --- judge-assessed severity (emitted only for \textsc{Divergent} and \textsc{Contradictory});
  \item \texttt{judge\_metadata} --- input/output token counts and decoding latency.
\end{enumerate}
The two narrative fields are the key methodological enabler of the downstream analyses described in §\ref{sec:characterization}. Clustering 34{,}706 \texttt{divergence\_topic} strings yields a 991-node taxonomy of disagreement themes; the \texttt{clinical\_significance} field permits stakes-adjusted aggregation. The judge prompt and the full output schema are in Appendix~\ref{app:judge_prompt}; inference is greedy (temperature~0) for reproducibility.

\paragraph{Comparison matrix.} For a question answered by $N$ handbooks the $\binom{N}{2}$ pairwise records and the integer matrix $\mathbf{M}\in\{0,\ldots,4\}^{N\times N}$ encoding the labels are written together as a single per-question JSON file. Diagonal entries are \textsc{Consistent} by convention. Per-question artefacts are independent and idempotent, enabling resume-safe incremental execution.

\subsection{Implementation and Scale}
\label{sec:scale}

The released pipeline runs over the full benchmark on a heterogeneous SLURM cluster (PSC Bridges-2, NVIDIA H100~80\,GB) with a sharded executor that splits the question set into 10 \emph{general} shards and 10 \emph{non-general} shards per pipeline stage; each shard is resumable at the matrix-file granularity for comparison and the question-file granularity for generation. The complete production run produces \textbf{48{,}056 answers} (Stage~2) and \textbf{5{,}730{,}465 pairwise comparisons} (Stage~3), of which 4{,}519{,}245 pre-screen as \textsc{Absent} and 1{,}211{,}220 require an LLM-judge call. Total wall-time and compute cost are reported in Appendix~\ref{app:compute}. To our knowledge this is the largest documented application of LLM-as-judge to a single medical heterogeneity benchmark.

\section{Validating the Evaluation Instrument}
\label{sec:validation}

The structured-output judge is the measurement instrument through which we read inter-source relationships; its trustworthiness underwrites every finding in §\ref{sec:characterization}. We validate it along two axes: agreement with human clinical annotators (\S\ref{sec:validation:human}) and an ablation against the natural alternative protocol --- a label-only judge followed by a post-hoc extractor --- confirming that the structured single-call design is required, not a convenience (\S\ref{sec:validation:ablation}).

\subsection{Human--judge agreement}
\label{sec:validation:human}

We validate the structured-output judge against human annotators on a stratified sample of 200 pairwise records (40 per non-absent label, plus 40 \textsc{Absent} controls); \textsc{Contradictory} is over-sampled at 46\% of all contradictions in the production run for power on the rare class. Annotators see the original question and both handbook answers; \textbf{the judge's label, reasoning, divergence topic, and clinical-significance rating are withheld}. Two annotators rate each pair following the operational definitions in Table~\ref{tab:taxonomy}; protocol and rubric in Appendix~\ref{app:annotation_protocol}.

\paragraph{Results.}
Both annotators completed all 200 pairs. Inter-annotator agreement is \textbf{Cohen's $\kappa = 0.655$} (raw agreement $73.0\%$) --- substantial under Landis--Koch. The two annotators agreed on 146/200 pairs; we treat their joint-agreed label as the human-majority gold. On those 146 pairs the judge agrees with the majority $87.7\%$ of the time, yielding \textbf{judge-vs-majority $\kappa = 0.842$} (almost perfect) and weighted F1 $= 0.876$ (macro F1 $= 0.841$). Per-label F1: \textsc{Absent} $1.00$, \textsc{Contradictory} $0.99$, \textsc{Consistent} $0.83$, \textsc{Complementary} $0.70$, \textsc{Divergent} $0.69$.

\paragraph{Failure-mode taxonomy.}
Of 18 judge errors against the majority, 14 (78\%) cluster on the \textsc{Complementary}/\textsc{Divergent} boundary: 8 cases where the majority calls \textsc{Complementary} but the judge calls \textsc{Divergent}, and 6 where the majority calls \textsc{Complementary} but the judge calls \textsc{Consistent}. The judge's discrimination is robust at the extremes (presence/absence; flat contradictions) but soft on the middle of the coverage--agreement axis --- consistent with the taxonomy's design intent that \textsc{Complementary} sits between \textsc{Consistent} and \textsc{Divergent}.

\paragraph{Clinical significance.}
On 49 paired \textsc{Divergent}/\textsc{Contradictory} pairs where all three (judge, A, B) rated significance, judge-vs-human $\kappa = 0.385$ --- fair but not strong. The judge's grades are directionally correct (no systematic \emph{low}/\emph{high} flips) but the fine-grained gradations should be treated as a population-level signal, not a per-pair adjudication.

\subsection{Structured vs.\ label-only judge: an ablation}
\label{sec:validation:ablation}

A natural alternative to our structured single-call judge is a label-only judge followed by a post-hoc extractor that conditions on (question, answer\textsubscript{a}, answer\textsubscript{b}, label) to recover \texttt{divergence\_topic} and \texttt{clinical\_significance} in a second call. We test the two protocols (Condition~A: structured single-call, ours; Condition~B: label-only + post-hoc) on the same 200-pair sample. Three findings emerge. (i)~Categorical agreement is $\kappa = 0.669$, but the disagreement concentrates on the most consequential class: of A's 40 \textsc{Divergent} pairs, B agrees on only 4 and downgrades 31 (78\%) to \textsc{Complementary}. (ii)~Clinical significance is unrecoverable post-hoc: on $n\!=\!44$ paired \textsc{Divergent}/\textsc{Contradictory} pairs B returns \emph{high} for all 44 ($\kappa\!=\!0$ against A's mixed \emph{high}/\emph{medium}). (iii)~Topic strings on agreed-label pairs are semantically equivalent and cluster identically under the \S\ref{sec:characterization} pipeline. Condition B is $\approx$$5$--$6\times$ faster per pair but loses the \textsc{Divergent}/\textsc{Complementary} discrimination and severity gradation. Structured single-call output is therefore a design requirement of the framework, not a convenience.

\section{Benchmark Characterization}
\label{sec:characterization}

We apply our pipeline to the full TransplantQA benchmark using Qwen3-32B as both generator and judge, reporting global and stratified label distributions, the per-organ heterogeneity profile, and a system-level comparison.

\subsection{Global Label Distribution}

Of the 5{,}730{,}465 pairwise comparisons, 4{,}519{,}245 (78.9\%) pre-screen as \textsc{Absent} because at least one handbook returned \texttt{NOT ADDRESSED}. Of the remaining 1{,}211{,}220 LLM-judged pairs, \textsc{Complementary} dominates (75.4\%), followed by \textsc{Divergent} (12.9\%), \textsc{Consistent} (7.1\%), and \textsc{Contradictory} ($<\!0.1\%$). Explicit contradiction is therefore rare; the dominant mode of disagreement is two centers covering different aspects of the same question (\textsc{Complementary}) or giving substantively different recommendations (\textsc{Divergent}).

\subsection{Per-Organ Heterogeneity}

Table~\ref{tab:per_organ} reports per-organ rates: the absence rate $r_{\mathrm{abs}}$, the per-pair divergence rate $R_{\mathrm{div}}$ (fraction of non-absent pairs labelled \textsc{Divergent} or \textsc{Contradictory}), the per-pair consistency rate $R_{\mathrm{con}}$, and the proportion of questions in each organ for which at least one pair is divergent ($\mathrm{pct}_{\mathrm{any\,div}}$).

\begin{table}[t]
\centering
\small
\caption{Per-organ heterogeneity rates from our reference production run. Per-pair rates are averaged over non-absent pairs.}
\label{tab:per_organ}
\begin{tabular}{l rrrr}
\toprule
 & $r_{\mathrm{abs}}$ & $R_{\mathrm{div}}$ & $R_{\mathrm{con}}$ & $\mathrm{pct}_{\mathrm{any\,div}}$ \\
\midrule
general  & 0.778 & 0.175 & 0.287 & 55.6\% \\
heart    & 0.692 & 0.146 & 0.177 & 45.3\% \\
kidney   & 0.655 & 0.157 & 0.161 & 49.5\% \\
liver    & 0.683 & 0.138 & 0.180 & 39.0\% \\
lung     & 0.596 & 0.148 & 0.156 & 54.2\% \\
pancreas & 0.668 & 0.185 & 0.202 & 29.9\% \\
\bottomrule
\end{tabular}
\end{table}

Absence dominates across all organs (60--78\%): even within the matching-organ subsets, the average handbook addresses only one third to half of relevant patient questions. Per-pair divergence rates cluster between 0.14 and 0.19, with pancreas and general questions sitting at the top of the range. The prevalence metric $\mathrm{pct}_{\mathrm{any\,div}}$ exhibits broader spread (30--56\%), reflecting that pancreas and liver questions are more often answered by a small subset of handbooks (so even when divergence exists, it concentrates within a few questions).

\subsection{Per-Handbook Coverage Spread}

Per-handbook absence rates span \textbf{0.45 to 0.99} (mean 0.74), a 2$\times$ spread between the most-comprehensive and most-silent handbooks. The handbook$\times$question-organ heatmap (Appendix Figure~\ref{fig:hb_by_organ_q}) shows the expected block-diagonal pattern but also systematic editorial differences: some handbooks are broadly comprehensive across all columns, while others are silent even within their own organ.

\subsection{System-Level Comparison: 14B Earlier Run vs.\ 32B Reference Run}
\label{sec:14B_vs_32B}

A previous run over the same benchmark used a hybrid-retrieval pipeline with Qwen3-14B as both generator and judge. Comparing it to the 32B reference run (per-organ deltas in Appendix~\ref{app:system_delta}) isolates the effect of the pipeline upgrade. Three observations stand out: (i)~absence drops 12--19~pp across every organ (mean $\Delta r_{\mathrm{abs}}=-0.136$) as better retrieval surfaces passages the earlier pipeline missed; (ii)~per-pair divergence rates are roughly unchanged or modestly lower (mean $\Delta R_{\mathrm{div}}=-0.031$; the stronger judge is not more aggressive); (iii)~the proportion of questions showing \emph{any} divergence rises substantially (mean $+15.9$~pp), driven mechanically by the absence drop. The per-pair rate reported by earlier baselines ($\approx20\%$) is thus stable, but the \emph{prevalence} of disagreement was substantially understated because absence was hiding it: stronger pipelines reveal latent disagreement rather than manufacturing it.

\subsection{Downstream Uses Enabled by Structured Output}

The two narrative fields support analyses that classifier-only judges cannot. Embedding the 16{,}113 unique \texttt{divergence\_topic} strings and clustering them yields a 991-theme taxonomy of \emph{what} sources disagree about (largest themes: post-transplant pregnancy timing, blood-test frequency, rejection symptoms, dental-care timing); the \texttt{clinical\_significance} field permits severity-weighted re-aggregation, which empirically tracks unweighted disagreement frequency closely (Spearman $\rho>0.99$ at the question, topic, and handbook levels) and is most useful for surfacing individual high-stakes pairs. These analyses are enabled by the structured judge output, not by the labels alone.

\section{Discussion}
\label{sec:discussion}

\paragraph{Generalisation to non-medical deployed RAG.}
The framework's three slots --- multi-source benchmark, inter-source taxonomy, structured-output judge --- are domain-agnostic. \emph{Legal RAG} (Westlaw AI, Lexis+ AI, Harvey) retrieves over jurisdictional layers and firm-specific research, yet single-gold benchmarks (LegalBench, LexGLUE) cannot surface whether a query grounded in California versus Texas precedent diverges in client-actionable ways. \emph{Educational RAG} retrieves over state-stratified standards (Common Core, NGSS) and publisher-specific expositions, while ScienceQA/GSM8K cannot surface whether a student's answer depends on which state's materials were indexed. Each instantiates the same slots with a domain-appropriate taxonomy: this paper's empirical contribution is medical, its methodological contribution is for deployed RAG generally.

\paragraph{Judge limitations.} An LLM judge inherits known biases \citep{zheng2024mtbench,kim2024prometheus}: \emph{self-preference} when generator and judge share a family (pair-symmetric framing mitigates but does not eliminate this; §\ref{sec:validation} measures $\kappa\!=\!0.842$ agreement), \emph{length/citation artefacts}, and \emph{cost} (Appendix~\ref{app:compute}).

\section{Conclusion}
\label{sec:conclusion}

We introduced TransplantQA, the HERO-QA retrieval system, and a structured-output LLM-as-judge as instruments for measuring inter-source heterogeneity in deployed medical RAG; all artefacts (48{,}056 answers, 5.73M pairwise comparisons, judge--majority $\kappa=0.842$) are released. Empirically, prior estimates understated the \emph{prevalence} of disagreement, not its intensity --- absence was hiding it. Methodologically, structured single-call judging is a requirement, not a convenience: post-hoc extraction loses the \textsc{Divergent}/\textsc{Complementary} discrimination and severity gradation the framework depends on.

\clearpage
\section*{Limitations}
\label{sec:limitations}

The empirical instantiation is confined to U.S.\ solid-organ transplant patient education (English, 2024--2025 snapshot); legal and educational transferability (\S\ref{sec:discussion}) is conceptual. The judge is an LLM; the 200-pair validation measures population-level agreement but cannot detect sub-axis biases (institution, organ, answer length) \citep{zheng2024mtbench,kim2024prometheus}; the released per-pair JSON preserves judge reasoning for individual-decision audit. Apparent inter-source divergence can also be inflated by retrieval failures rather than true disagreement; the absence pre-screen partially mitigates this.

\bibliography{custom}

\clearpage
\onecolumn
\appendix

\section{Question sources and inclusion criteria}\label{app:question_sources}

The 1{,}115 released questions were drawn from an initial pool of 3{,}000+ candidates collected from four families of public, patient-facing sources. Table~\ref{tab:question_sources} reports the top-10 source names in the final benchmark.

\begin{table}[H]
\centering
\small
\caption{Top-10 source names for the released question set, by number of contributing questions.}
\label{tab:question_sources}
\begin{tabular}{lr}
\toprule
\textbf{Source} & \textbf{N questions} \\
\midrule
Mayo Clinic patient forums          & 280 \\
Reddit (transplant subreddits)      & 268 \\
National Kidney Foundation Q\&A     & 162 \\
American Liver Foundation Q\&A      &  93 \\
Healthy Transplant                  &  29 \\
Inspire (community)                 &  21 \\
HRSA (US Government)                &  15 \\
RWJBarnabas Health                  &  14 \\
Endocrine Society                   &  13 \\
UCSF Health                         &  12 \\
\bottomrule
\end{tabular}
\end{table}

Source families (final shares): institutional Q\&A pages (31.2\%), community forums such as Reddit and Mayo Clinic Connect (25.1\%), patient-facing medical organizations (24.9\%), and a long tail of government health agencies and patient advocacy sites (18.8\%). 69.9\% of questions are geolocated to the United States.

\textbf{Collection and inclusion.} Candidate questions were harvested from the source platforms above using transplant- and symptom-keyword search. A candidate was retained if it (a)~was \emph{relevant} to transplant patient education (excluding administrative or off-topic questions) and (b)~was \emph{non-duplicative} of an earlier-retained question (cosine deduplication at threshold 0.85 followed by manual review of near-duplicates). Every retained question was then \emph{anonymized and rephrased} to (c)~strip personally identifying information about the asker or named individuals and (d)~make the question \emph{self-contained} (interpretable without surrounding conversational context).

\section{Topic taxonomy}\label{app:topic_taxonomy}

Each question is annotated with one or more of 13 top-level topic categories. Table~\ref{tab:topic_taxonomy} lists the categories and their share of the question set (multi-label, percentages can sum to $>$100\%).

\begin{table}[H]
\centering
\small
\caption{13-topic taxonomy for the question set. Multi-label.}
\label{tab:topic_taxonomy}
\begin{tabular}{lr}
\toprule
\textbf{Topic} & \textbf{Share (\%)} \\
\midrule
Medical Complications                & 28.6 \\
Reproductive Health                  & 26.1 \\
Lifestyle \& Daily Living            & 20.2 \\
Pre-Transplant                       & 15.8 \\
Medications                          &  9.9 \\
Monitoring \& Follow-up              &  9.7 \\
Mental \& Emotional Health           &  9.6 \\
Surgery \& Recovery                  &  7.0 \\
Special Populations \& Education     &  6.5 \\
Transplant Process \& Logistics      &  1.9 \\
Financial \& Insurance               &  0.6 \\
Financial \& Administrative          &  1.7 \\
Support \& Community                 &  0.5 \\
\bottomrule
\end{tabular}
\end{table}

A second tier of 43 fine-grained sub-topic tags refines these categories (e.g., \emph{Medications $\rightarrow$ Tacrolimus interactions}; \emph{Reproductive Health $\rightarrow$ Mycophenolate timing before pregnancy}). The sub-topic list is included in the released annotation file.

\section{Data card (Datasheet for Datasets)}\label{app:datacard}

Following the recommendations of \citet{gebru2021datasheets}, we provide a structured data card.

\textbf{Motivation.} Created to enable evaluation of medical RAG systems on a corpus with genuine institutional heterogeneity, and to enable analysis of that heterogeneity itself.

\textbf{Composition.} 1{,}115 patient-derived questions; 102 transplant patient-education handbooks from 23 U.S.\ centres across 5 organ types; 48{,}056 grounded answers from the reference production run; 5{,}730{,}465 pairwise comparisons (1{,}211{,}220 LLM-judged, 4{,}519{,}245 absence-pre-screened); per-question matrices; per-shard summaries.

\textbf{Collection.} Questions collected over [date range] from public sources listed in Appendix~\ref{app:question_sources}. Handbooks downloaded as PDFs from public institutional websites in 2024--2025. No interaction with patients or clinicians for data collection.

\textbf{Preprocessing.} Questions lightly paraphrased for anonymisation and self-containment. Handbooks extracted from PDF via LlamaParse; chunked at section boundaries with 512-token sub-chunking. Answers and judgments produced by Qwen3-32B at temperature 0.

\textbf{Uses.} Intended for evaluating medical RAG systems' behaviour under multi-source corpora, for measuring institutional heterogeneity in patient education, and as a benchmark for new LLM judges. \textbf{Not intended} for ranking individual transplant centres or for direct clinical decision support.

\textbf{Distribution.} Release-location metadata is anonymized for review. Original handbook PDFs are not redistributed but are listed by URL.

\textbf{Maintenance.} Maintained by the authors, with annual updates planned when new handbook revisions are detected.

\section{Answer-generation prompt}\label{app:gen_prompt}

The reference production run uses the HERO-QA system prompt (the \texttt{HERO\_QA\_SYSTEM\_PROMPT} below) paired with the \texttt{USER\_TEMPLATE} for evidence framing. The earlier hybrid-retrieval baseline used a comparable system prompt without the section-citation requirement.

\begin{quote}\small\ttfamily
System: You are a clinical information assistant using HERO-QA evidence. Answer the patient's question based ONLY on the provided handbook evidence from this specific transplant center. Follow these rules strictly:\\
1. If the evidence answers the question, give the answer using only that evidence.\\
2. Cite the supporting section heading, and page if provided. If pages are unknown, cite the section heading only.\\
3. If the evidence does not answer the question, respond exactly: "NOT ADDRESSED: This handbook does not contain information on this topic."\\
4. Do not use outside medical knowledge. Do not fill gaps with general transplant advice.\\[4pt]
User: \#\# Handbook Context\\
\{context\}\\[2pt]
\#\# Patient Question\\
\{question\}
\end{quote}

Generation runs with greedy decoding (temperature~0), \texttt{max\_new\_tokens=512}, and \texttt{<think>...</think>} reasoning blocks stripped before the answer is persisted.

\section{Judge prompt and output schema}\label{app:judge_prompt}

Our judge uses two prompts: a binary absence-detection prompt (used only when the heuristic \texttt{NOT ADDRESSED} prefix is not detected) and the main comparison prompt.

\paragraph{Absence-detection prompt.}
\begin{quote}\small\ttfamily
You are a clinical information assistant. Read the following response that was generated from a transplant center handbook and determine whether it effectively states that the handbook does NOT contain information on the topic.\\[4pt]
Response:\\
\{answer\}\\[4pt]
Does this response indicate the handbook does not address the question? Answer with exactly one word: YES or NO
\end{quote}

\paragraph{Comparison prompt.}
\begin{quote}\small\ttfamily
You are a clinical expert evaluating whether two transplant center handbooks give consistent guidance on the same patient question.\\[4pt]
\#\# Task\\
Compare Answer A and Answer B and classify their relationship as exactly one of:\\
ABSENT / CONSISTENT / COMPLEMENTARY / DIVERGENT / CONTRADICTORY\\[4pt]
\#\# Definitions\\
- ABSENT: One or both answers indicate the handbook does not contain information on the topic, so no meaningful comparison can be made.\\
- CONSISTENT: Both answers provide substantive clinical content and give the same clinical recommendation.\\
- COMPLEMENTARY: Both answers provide substantive clinical content that is compatible, but they differ in level of detail.\\
- DIVERGENT: Both answers provide substantive clinical content but differ in a clinically meaningful way (different thresholds, timelines, or recommendations that would lead to different patient behavior).\\
- CONTRADICTORY: Both answers provide substantive clinical content that gives directly opposing guidance.\\[4pt]
IMPORTANT: If either answer states the handbook does not address the topic, or provides no substantive clinical content, you MUST classify the pair as ABSENT.\\[4pt]
\#\# Input\\
Question: \{question\}\\
Answer A (\{center\_a\}): \{answer\_a\}\\
Answer B (\{center\_b\}): \{answer\_b\}\\[4pt]
\#\# Output (JSON only, no other text)\\
\{\{\\
\ \ "classification": "<label>",\\
\ \ "reasoning": "<2-3 sentence clinical justification>",\\
\ \ "divergence\_topic": "<specific sub-topic of divergence, if applicable, else null>",\\
\ \ "clinical\_significance": "<low/medium/high if divergent or contradictory, else null>"\\
\}\}
\end{quote}

\paragraph{Output schema and parsing.} Judge outputs are parsed as JSON; if parsing fails, a fallback extractor scans the raw text for a recognised label and assigns the remaining fields to \texttt{null}. Across the 1{,}211{,}220 LLM-judged pairs in the reference run, JSON parsing succeeded on $>$99.5\% of calls.

\section{Compute cost}\label{app:compute}

Production runs used NVIDIA H100 80\,GB GPUs on PSC Bridges-2 via a SLURM allocation. Wall-time figures aggregate the 20 generation shards and 20 comparison shards from the released reference run.

\begin{table}[H]
\centering
\small
\caption{Approximate compute cost of the reference production run.}
\label{tab:compute}
\begin{tabular}{lrr}
\toprule
\textbf{Stage} & \textbf{Wall hours} & \textbf{GPU-hours (H100-80)} \\
\midrule
Document extraction (CPU) & 12   & 0 \\
Indexing                  & 8    & 8 \\
Stage 2: Generation       & $\approx$130 & $\approx$130 \\
Stage 3: Comparison       & $\approx$300 & $\approx$300 \\
\midrule
\textbf{Total}            & $\approx$450 & $\approx$438 \\
\bottomrule
\end{tabular}
\end{table}

At an indicative H100-80 GB cloud rate of \$3--4/hour, the total reference-run cost is approximately \$1.3K--\$1.8K. The pipeline is fully resumable: a stalled or pre-empted shard can be re-launched without recomputing its already-persisted per-question artefacts. Smaller domains (10--20 handbooks) are runnable on a single H100 in under 24~hours.

\clearpage
\section{Per-handbook coverage heatmap}\label{app:coverage_heatmap}

\begin{figure}[H]
\centering
\includegraphics[height=0.8\textheight]{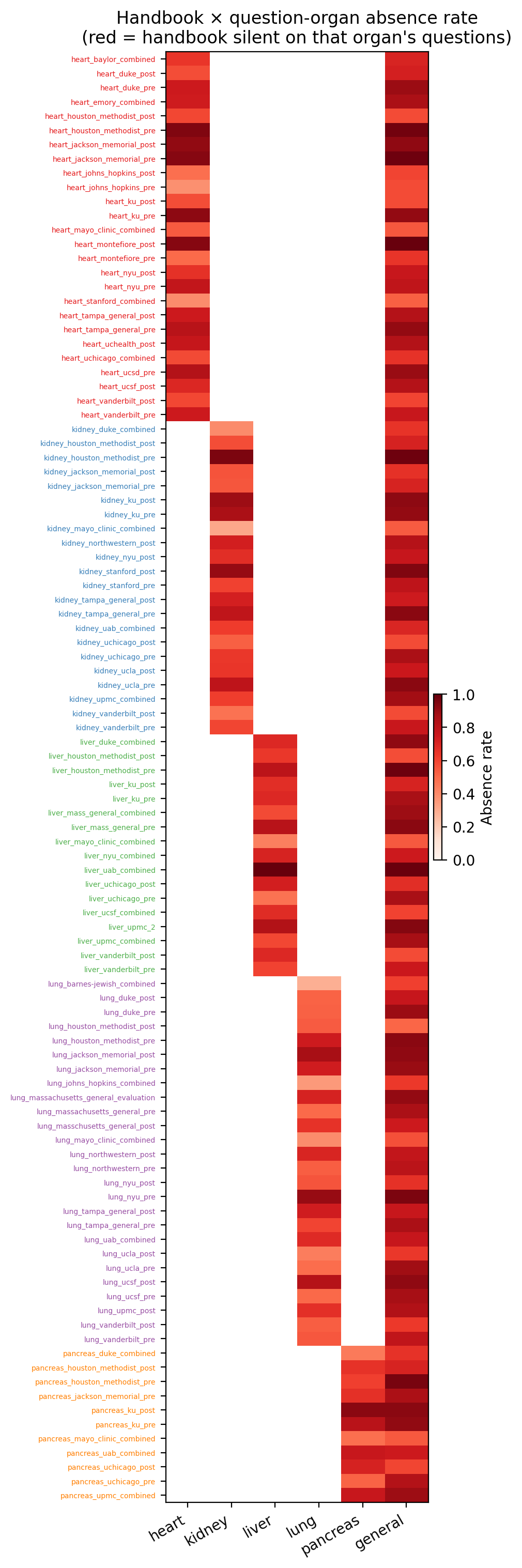}
\caption{Handbook $\times$ question-organ absence rate. Rows are the 102 handbooks (grouped and colour-coded by organ); columns are the six question-organ groups. Red = the handbook is silent on that organ's questions. The block-diagonal structure reflects that organ-specific handbooks answer mainly their own-organ and general questions; rows that are pale across all columns (e.g.,\ several Mayo Clinic, UChicago, Houston Methodist handbooks) are broadly comprehensive.}
\label{fig:hb_by_organ_q}
\end{figure}

\clearpage
\section{System-level delta, 14B vs.\ 32B}\label{app:system_delta}

Table~\ref{tab:14b_vs_32b} reports the per-organ deltas underlying the system-level comparison in \S\ref{sec:14B_vs_32B}.

\begin{table}[H]
\centering
\small
\caption{System delta: 32B reference run (HERO-QA retrieval + 32B judge) $-$ 14B earlier run (hybrid retrieval + 14B judge). The pipeline upgrade systematically lowers absence without inflating per-pair divergence; instead, the \emph{prevalence} of divergence rises.}
\label{tab:14b_vs_32b}
\begin{tabular}{l rrrr}
\toprule
 & $\Delta r_{\mathrm{abs}}$ & $\Delta R_{\mathrm{div}}$ & $\Delta R_{\mathrm{con}}$ & $\Delta \mathrm{pct}_{\mathrm{any\,div}}$ \\
\midrule
general  & $-0.127$ & $-0.014$ & $+0.002$ & $+0.270$ \\
heart    & $-0.126$ & $-0.006$ & $-0.010$ & $+0.190$ \\
kidney   & $-0.125$ & $-0.080$ & $+0.014$ & $+0.097$ \\
liver    & $-0.122$ & $-0.071$ & $+0.037$ & $+0.085$ \\
lung     & $-0.129$ & $-0.003$ & $+0.020$ & $+0.124$ \\
pancreas & $-0.188$ & $-0.010$ & $-0.020$ & $+0.188$ \\
\midrule
\textbf{mean} & $-0.136$ & $-0.031$ & $+0.007$ & $+0.159$ \\
\bottomrule
\end{tabular}
\end{table}

\section{Annotation protocol}\label{app:annotation_protocol}

The full validation protocol --- sample design, annotator-facing rubric with operational tiebreakers, clinical-significance definitions, calibration plan, quality assurance, and scoring metrics --- is provided as supplementary material under \texttt{drafts/annotation\_study/PROTOCOL.md}. The 200-pair stratified sample (\texttt{sample\_v1/annotation\_sample\_full.csv}), two shuffled annotator-facing packets (\texttt{packets/annotator\_\{A,B\}.csv}), and the deterministic sampler (\texttt{src/analysis/build\_annotation\_sample.py}) are released alongside the benchmark for full reproducibility.

\end{document}